\documentclass{article}
\usepackage{spconf,amsmath,graphicx,hyperref}
\usepackage{verbatim}
\usepackage{csquotes}
\usepackage{multirow}
\usepackage{graphicx}
\usepackage{booktabs}
\usepackage{pifont}
\newcommand{\cmark}{\ding{51}}
\newcommand{\xmark}{\ding{55}} 

\title{AVSD-Scenes: A Dataset for Audio-Visual Description of Urban  Scenes}
\name{Dhanunjaya Varma Devalraju$^{1,*}$, Arshdeep Singh$^{2,*}$, Mark D. Plumbley$^{2}$\thanks{This work was supported by the Engineering and Physical Sciences Research Council (EPSRC)  [grant number EP/Y028805/1]. For the purpose of open access, the authors have applied a Creative Commons Attribution (CC BY) licence to any Author Accepted Manuscript version arising.} } 

\address{ $^{1}$IIT Madras, India, $^{2}$King's College London, United Kingdom }
\begin{document}
\ninept
\maketitle
\def\thefootnote{*}\footnotetext{These authors contributed equally to this work}\def\thefootnote{\arabic{footnote}}
\begin{abstract}

Natural language descriptions can provide rich semantic representations of audio-visual urban scenes, yet datasets that jointly describe both auditory and visual information remain limited. In this paper, we introduce AVSD-Scenes, a paired audio-visual scene description dataset for urban environments. The dataset contains 12,291 audio-visual scene descriptions generated from the TAU Urban Audio-Visual Scenes dataset. To construct the dataset, we first generate audio- and visual-based descriptions using Qwen2-Audio-7B and Qwen2.5-VL-7B, respectively. These modality-specific descriptions are then combined using large language models, namely Qwen3-14B, Mistral-Small-3.2-24B-Instruct-2506, and Gemma-3-27B-it, to produce multimodal descriptions that capture complementary information from both modalities. We benchmark AVSD-Scenes using semantic alignment, cross-modal retrieval, scene classification, LLM-as-a-judge evaluation, and human subjective assessment. Results show that multimodal descriptions improve semantic alignment and cross-modal retrieval performance compared with modality-specific descriptions while preserving strong scene-discriminative information. The generated descriptions achieve up to 94.5\% accuracy in urban scene classification, while combining audio, visual, and description embeddings further improves accuracy to 95.4\%. Furthermore, the descriptions remain highly scene-discriminative even when scene labels are removed from the prompting instructions, indicating that they capture semantic information derived from the audio-visual content rather than merely reflecting label information.  

\end{abstract}
\section{Introduction}
\label{sec:intro}

Urban scenes contain diverse auditory and visual cues that jointly characterize the surrounding environment. Audio conveys activities and events, while visual information provides context about the scene and potential sound sources. Natural-language descriptions of urban scenes provide rich semantic context about the environment, activities, and audio-visual events beyond conventional scene labels.  Such descriptions are valuable for tasks including scene understanding \cite{liu2023open}, retrieval \cite{wang2025cross}, and multimodal language generation \cite{wu2023multimodal}.

Audio captioning \cite{mei2022automated} aims to bridge acoustic signals and natural language by generating textual descriptions of audio content. 
Earlier advances in audio captioning were facilitated by human annotated datasets such as Clotho \cite{drossos2020clotho}, a widely used benchmark dataset  comprising  approximately 5k environmental audio clips collected from Freesound\footnote{https://freesound.org/} and  paired with multiple human-written captions. Subsequently, AudioCaps \cite{kim2019audiocaps} introduced a larger benchmark with approximately 46k audio-text pairs collected from AudioSet \cite{gemmeke2017audioset}.  However, the cost and effort required for manual annotation limits the scalability of such datasets \cite{bach2019snorkel}.  To address this challenge, subsequent studies have explored automated approaches for constructing large-scale audio-language datasets from publicly available resources such as BBC sound effects\footnote{https://sound-effects.bbcrewind.co.uk/}, Freesound and the AudioSet. These audio captioning datasets include LAION-Audio-630K \cite{wu2023large}, WavCaps \cite{mei2024wavcaps} and AudioSetCaps \cite{bai2025audiosetcaps}, and have captions generated  from audio content using large language models. Other datasets, such as Auto-ACD \cite{sun2024auto} and Sound-VECaps \cite{yuan2025sound} are built on audio-visual datasets such as AudioSet \cite{gemmeke2017audioset} and VGG-Sound \cite{chen2020vggsound}, rely on automated caption generation pipelines that combine audio, visual, and language models to create  audio-text pairs.

\begin{figure}[t]
    \centering
    \includegraphics[width=1\linewidth]{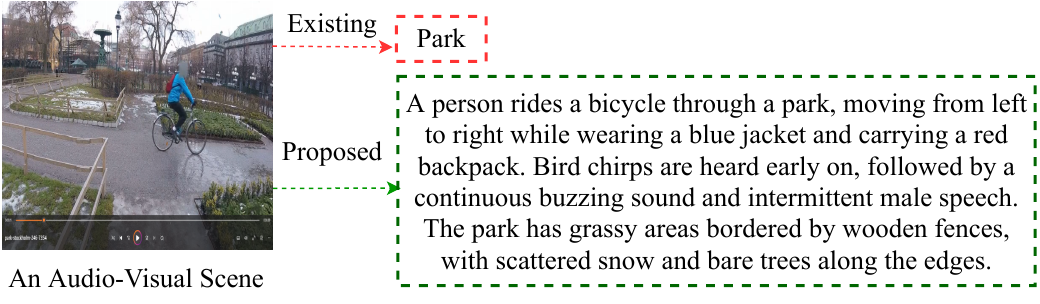}
    \caption{Comparison of AVSD-Scenes descriptions with existing descriptions of an audio-visual scene example \enquote{park-stockholm-246-7354} taken from  TAU Urban Audio-Visual Scenes 2021 development set ~\cite{wang2021curated}.}
    \label{fig: comaprison for intro}
\end{figure}

Audio captioning datasets derived from AudioSet \cite{gemmeke2017audioset} primarily describe acoustic events, while audio-visual datasets such as VGG-Sound \cite{chen2020vggsound} provide event-level annotations associated with visible sound sources. In contrast, the TAU Urban Audio-Visual Scenes dataset \cite{wang2021curated} captures holistic urban environments using standardized recording protocols, providing more consistent multimodal data than earlier large-scale audio-visual datasets \cite{abu2016youtube,caba2015activitynet,wang2020mead}. However, the TAU dataset represents each environment using only predefined scene labels, such as \enquote{airport} or \enquote{park}. These coarse labels omit contextual details, activities, and audio-visual interactions, limiting their ability to represent real-world urban scenes through rich natural-language descriptions.



To address this limitation, we introduce AVSD-Scenes, an Audio-Visual Scene Description dataset for urban acoustic scenes. AVSD-Scenes contains natural language descriptions generated from both audio and visual scene information. Unlike existing  audio-visual scene dataset \cite{wang2021curated} that rely on predefined scene categories, AVSD-Scenes provides detailed audio-visual scene descriptions that capture the overall context of a scene in natural language as shown in Fig. \ref{fig: comaprison for intro}. By jointly leveraging audio and visual information, the proposed dataset offers a richer representation of real-world urban environments compared to existing audio-visual scene dataset \cite{wang2021curated}, and has the potential to support future research on audio-visual scene understanding, retrieval, and multimodal language generation.

The main contributions of this work are as follows: 

\begin{enumerate}
\item We introduce \textbf{AVSD-Scenes}, a dataset of 12,291 paired audio-visual scene descriptions, extending an existing urban audio-visual scene dataset \cite{wang2021curated} with natural language descriptions.

\item We propose an automated pipeline for generating multimodal scene descriptions by fusing audio- and visual-specific descriptions generated by foundation models.

\item We establish benchmarks for cross-modal retrieval, semantic alignment, scene classification, and description quality evaluation using both LLM-based and human assessment.

\item We show that multimodal descriptions improve retrieval and semantic alignment over modality-specific descriptions, retain scene-discriminative information even without scene-label information in the instructions, and provide complementary information beyond conventional audio-visual representations.
The dataset, code, and evaluation criterion are publicly available on Github.\footnote{\label{fn:github}\url{https://github.com/dhanunjaya-varma/AVSD-Scenes.git}}

\end{enumerate}

\section{Dataset construction pipeline}
\label{sec:related}



We construct AVSD-Scenes, an audio-visual scene description dataset derived from the TAU Urban Audio-Visual Scenes 2021 development dataset \cite{wang2021curated}, which  is collected in 10 European cities, where each sample consists of an audio recording accompanied by its corresponding visual sequence. The original dataset provides synchronized audio-visual recordings with scene-level labels.

In this work, we augment these audio-visual samples with natural-language descriptions that, beyond the predefined scene labels, explain the acoustic and visual events occurring in the recorded scenes, thereby providing richer semantic information about the underlying audio-visual content. 

\begin{figure}[t]
    \centering
    \includegraphics[width=1\columnwidth]{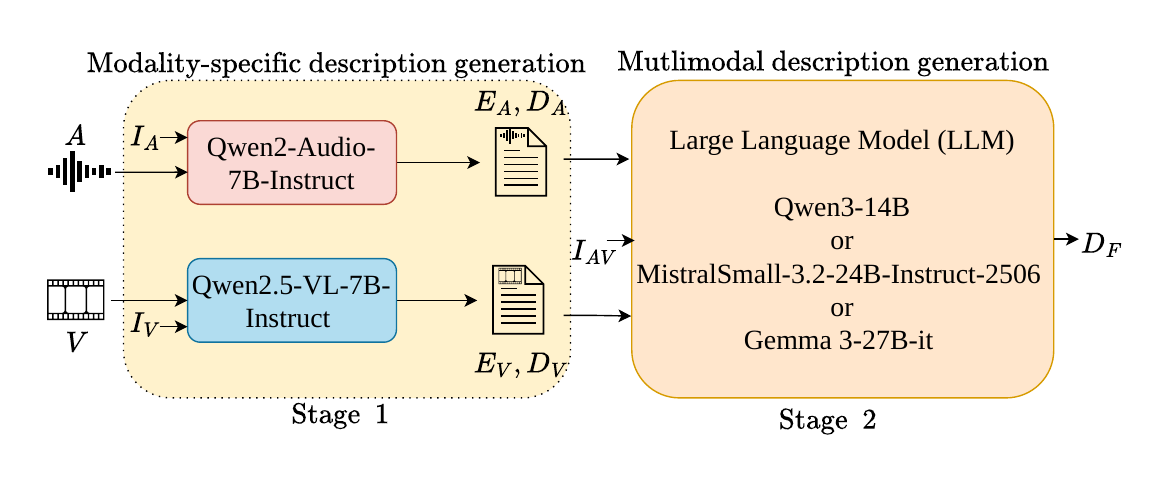}
    \vspace{-0.95cm}
    \caption{Dataset construction pipeline.  All the models are loaded using 4-bit NormalFloat quantization with double quantization and FP16 computation for efficient inference.}
    \label{fig: overall pipeline}
\end{figure}

As shown in Fig.~\ref{fig: overall pipeline}, the dataset is constructed in two stages. In the first stage, the audio and visual recordings are processed independently to identify acoustic and visual events, respectively, and to generate modality-specific descriptions. In the second stage, these modality-specific events and descriptions are jointly used to generate a unified scene description using a large language model (LLM).


\subsection{Stage 1: Modality-specific description generation}
\label{stage1}
We use the instruction-tuned models \textit{Qwen2-Audio-7B-Instruct} \cite{chu2024qwen2} and \textit{Qwen2.5-VL-7B-Instruct} \cite{bai2025qwen2_5vl} to identify modality-specific events and generate corresponding descriptions.  For the audio modality, \textit{Qwen2-Audio-7B-Instruct} is instructed ($I_A$) to identify sound events in the recording, describe them in chronological order, including overlapping events when applicable, and generate a concise caption summarizing the recording. 


For the visual modality, \textit{Qwen2.5-VL-7B-Instruct} is instructed ($I_V$) to identify visible objects, people, animals, vehicles, and actions, describe them chronologically, including simultaneous actions when applicable, and generate a concise caption summarizing the visual content. The model is instructed to avoid inferring unobservable intentions or events and to describe visual characteristics when an object or action cannot be confidently identified. For both models, the scene label is provided as contextual information inside the instructions.

Given an audio recording $A$, visual sequence $V$, instructions $I_A/I_V$, we generate the sound events $E_A$, audio descriptions $D_A$, visual events $E_V$, and visual descriptions $D_V$ as follows:

\begin{equation}
(E_A,D_A) = f_{\mathrm{Qwen2\text{-}Audio}}(A, I_A),
\end{equation}
\begin{equation}
(E_V,D_V) = f_{\mathrm{Qwen2.5\text{-}VL}}(V, I_V).
\end{equation}


\subsection{Stage 2: Multimodal descriptions generation}
We use LLMs to generate a unified scene description from the modality-specific events and descriptions generated in Stage 1. Specifically, we use \textit{Qwen3-14B} \cite{yang2025qwen3} (denoted as \enquote{Qwen3}), \textit{Mistral-Small-3.2-24B-Instruct-2506}\cite{mistralsmall2025} (denoted as \enquote{Mistral}), and \textit{Gemma-3-27B-it} \cite{team2025gemma} (denoted as \enquote{Gemma}) as fusion models.  Given the audio events $E_A$, audio description $D_A$, visual events $E_V$, visual description $D_V$ and, instructions $I_{AV}$, the final description $D_F$ is generated as
\begin{equation}
D_F =
f_{\mathrm{LLM}}(E_A,D_A,E_V,D_V, I_{AV}).
\end{equation}

Each model is instructed ($I_{AV}$) to combine relevant information from both modalities while ensuring that the generated description is grounded in the provided events. The instruction also emphasizes avoiding unsupported inferences, hallucinated details, and redundancy.  The scene label  is  also provided as contextual information inside the instruction. The final description is limited to approximately $30$ - $60$ words and presented as a single paragraph.
Additional implementation details, prompts/instructions, and dataset statistics are provided on GitHub$^{\ref{fn:github}}$.



\section{Evaluation framework}




\subsection{Semantic alignment and cross-modal retrieval benchmark} 

To assess how well multimodal representations capture the shared semantic content between audio, visual, and textual modalities, we  benchmark AVSD-Scenes by measuring semantic alignment and cross-modal scene-level  retrieval performance between audio-visual clips and their corresponding scene descriptions.

We measure semantic alignment between embeddings generated from audio, visual or text modality. For text modality, we use either modality-specific or multimodal description. We evaluate scene-level retrieval tasks to analyse, \textit{\enquote{Can our generated descriptions retrieve an audio/visual belonging to the correct scene}}. In the text-to-audio/visual retrieval task,  the modality-specific or multimodal descriptions are  used as a query to retrieve the corresponding audio/visual recording from the dataset.

Given audio or visual clips $\mathcal{X}=\{x_1,\ldots,x_N\}$ and scene descriptions $\mathcal{T}=\{t_1,\ldots,t_N\}$, pretrained multimodal models project both modalities into a shared embedding space. Their similarity is computed using cosine similarity:

\begin{equation} s(x_i,t_j)=\frac{\mathbf{z}_{x_i}^{\top}\mathbf{z}_{t_j}} {\|\mathbf{z}_{x_i}\|\|\mathbf{z}_{t_j}\|} \end{equation} 

 
\noindent where $\mathbf{z}_{x_i}$ and $\mathbf{z}_{t_j}$ denote the embeddings of clip $x_i$ and description $t_j$, respectively.






We report recall at rank $k$ ($\mathrm{R@k}$). For each query, $\mathrm{R@k}=1$ if the correct item appears among the top-$k$ retrieved results, and $0$ otherwise. The final score is averaged over all queries.

\noindent \textbf{Multimodal embedding models:}
We use three pretrained models, CLIP~\cite{radford2021learning}, CLAP~\cite{elizalde2023clap}, and IMAGEBIND~\cite{girdhar2023imagebind}, to extract embeddings for cross-modal retrieval and semantic similarity evaluation. CLIP learns a joint image--text representation by aligning paired images and natural-language descriptions through contrastive learning, while CLAP similarly learns a joint audio--text representation from paired audio and text. IMAGEBIND learns a shared embedding space across multiple modalities, including text, audio, and vision.







\subsection{Scene classification evaluation}
To assess whether the generated multimodal descriptions preserve scene-discriminative information, we evaluate their effectiveness for downstream scene classification using the TAU Urban Audio--Visual Scenes 2021 development dataset~\cite{wang2021curated}. We follow the predefined split, with  70\% of the data used for training and 30\% for testing.

\noindent \textbf{Feature representation:} For each example, we extract modality-specific embeddings from the audio, visual, and our generated scene descriptions. Following~\cite{wang2021curated}, we use audio and visual embeddings extracted using the pre-trained $L^{3}$-Net~\cite{l3net}, resulting in $512$-dimensional representations for each modality. The scene descriptions are encoded using the pre-trained \texttt{bert-base-uncased} model~\cite{devlin2019bert}, and the $768$-dimensional representation of the \texttt{[CLS]} token is used as the textual embedding. Further, we employ an early feature fusion strategy, in which embeddings from different modalities are concatenated to form a joint feature  vector for classification.\\ 
\noindent \textbf{Classification:} We use a simple one-vs-rest support vector machine (SVM) with radial basis function kernel with regularization parameter of 10 and train the SVM using the training dataset and evaluate it on test dataset.



\subsection{Description quality evaluation}

We evaluate description quality using both an LLM-as-a-judge framework and human subjective assessment. 


\noindent \textbf{LLM-as-a-judge evaluation:} 
We use an LLM-as-a-judge framework~\cite{gu2026survey} with Qwen2.5-14B-Instruct~\cite{chu2024qwen2} to evaluate the generated audio--visual scene descriptions. Each description is rated on a five-point scale across eight criteria: audio fidelity, visual fidelity, event coverage, hallucination, cross-modal consistency, grammar, fluency, and overall quality. For hallucination, higher scores indicate fewer unsupported details. Scores are averaged across the dataset to obtain an overall assessment for each model.


\noindent \textbf{Human evaluation:} 
To complement the automatic evaluation, we conduct a human subjective evaluation on 100 randomly selected audio-visual samples balanced across scene classes, following ethical approval. Four participants independently rated the Mistral-generated multimodal descriptions on a five-point scale using six criteria: audio fidelity, visual fidelity, event coverage, hallucination, fluency, and overall quality.

\begin{table}[t]
\centering
\caption{Cross-modal similarity of the generated descriptions. Avg. denotes the mean of the text--visual and text--audio similarity.}
\LARGE
\label{tab:similarity_scores}
\setlength{\tabcolsep}{3.2pt}
\renewcommand{\arraystretch}{1.05}
\resizebox{1\columnwidth}{!}{%
\begin{tabular}{lcc|c|cc|c}
\toprule
\multirow{2}{*}{Description}
& \multicolumn{3}{c|}{CLIP / CLAP}
& \multicolumn{3}{c}{IMAGEBIND} \\
& Text--Visual & Text--Audio & Avg.
& Text--Visual & Text--Audio & Avg. \\
\midrule
Modality-Specific
& .3299 & .2143 & .2721
& .3627 & .0556 & .2092 \\
\midrule
Multimodal (Qwen3)
& .3187 & .2854 & .3021
& .3425 & .1696 & .2561 \\
Multimodal (Mistral)
& .3018 & .2613 & .2816
& .3165 & .1615 & .2390 \\
Multimodal (Gemma)
& .3083 & .2735 & .2909
& .3218 & .1599 & .2409 \\
\bottomrule
\end{tabular}%
}
\end{table}

\begin{table*}[t]
\centering
\caption{Scene-level cross-modal retrieval performance (\%). Avg. denotes
the mean of text--audio (T$\rightarrow$A) and text--visual
(T$\rightarrow$V) retrieval.}
\label{tab:scene_retrieval}
\Large
\setlength{\tabcolsep}{2.5pt}
\renewcommand{\arraystretch}{1.05}
\resizebox{0.95\textwidth}{!}{%
\begin{tabular}{lccc|ccc|ccc|ccc|ccc|ccc}
\toprule
\multirow{3}{*}{\textbf{Description}}
& \multicolumn{9}{c|}{\textbf{CLAP / CLIP}}
& \multicolumn{9}{c}{\textbf{IMAGEBIND}} \\

& \multicolumn{3}{c}{\textbf{T$\rightarrow$A}}
& \multicolumn{3}{c}{\textbf{T$\rightarrow$V}}
& \multicolumn{3}{c|}{\textbf{Avg.}}
& \multicolumn{3}{c}{\textbf{T$\rightarrow$A}}
& \multicolumn{3}{c}{\textbf{T$\rightarrow$V}}
& \multicolumn{3}{c}{\textbf{Avg.}} \\

& R@1 & R@5 & R@10
& R@1 & R@5 & R@10
& R@1 & R@5 & R@10
& R@1 & R@5 & R@10
& R@1 & R@5 & R@10
& R@1 & R@5 & R@10 \\
\midrule

Modality-Specific
& 24.99 & 49.97 & 60.85
& 85.19 & 94.57 & 97.13
& 55.09 & 72.27 & 78.99
& 24.40 & 40.95 & 51.82
& 90.76 & 97.01 & 98.39
& 57.58 & 68.98 & 75.11 \\

\midrule

Multimodal (Qwen3)
& 41.43 & 68.00 & 77.06
& 84.49 & 93.98 & 96.70
& 62.96 & 80.99 & 86.88
& 58.55 & 82.12 & 89.87
& 90.56 & 96.89 & 98.23
& 74.56 & 89.51 & 94.05 \\

Multimodal (Mistral)
& 37.00 & 63.82 & 73.20
& 85.47 & 94.91 & 97.35
& 61.23 & 79.36 & 85.27
& 59.78 & 82.52 & 90.23
& 89.57 & 96.72 & 98.23
& 74.68 & 89.62 & 94.23 \\

Multimodal (Gemma)
& 38.35 & 65.74 & 75.40
& 83.04 & 93.50 & 96.36
& 60.69 & 79.62 & 85.88
& 56.46 & 78.66 & 86.05
& 87.78 & 95.17 & 97.25
& 72.12 & 86.92 & 91.65 \\

\bottomrule
\end{tabular}%
}
\end{table*}

\section{Dataset benchmarking and analysis}



Table~\ref{tab:similarity_scores} presents the similarity between the generated descriptions and the audio and visual modalities using CLIP/CLAP and IMAGEBIND models. The modality-specific descriptions obtain higher text--visual similarity, while the multimodal descriptions improve text--audio similarity and maintain similar text--visual similarity.   Among the multimodal descriptions, Qwen3 obtains the highest average similarity for both evaluations. These results show that combining audio and visual information in the descriptions increases their correspondence with the audio while retaining most of their correspondence with the visual content.

Table~\ref{tab:scene_retrieval} also shows  similar findings for cross-modal scene retrieval. The multimodal descriptions improve text-to-audio retrieval with both CLAP and IMAGEBIND compared with the modality-specific descriptions. On the other hand, text-to-visual retrieval performance remains high and similar for both modality-specific and multimodal descriptions.
Overall, the similarity and retrieval results show that combining information from audio and visual modalities improves the representation of audio content in the generated descriptions while retaining most of the visual information  as captured by the modality-specific descriptions.


\begin{figure}[ht]
    \centering
    \includegraphics[width=0.981\columnwidth]{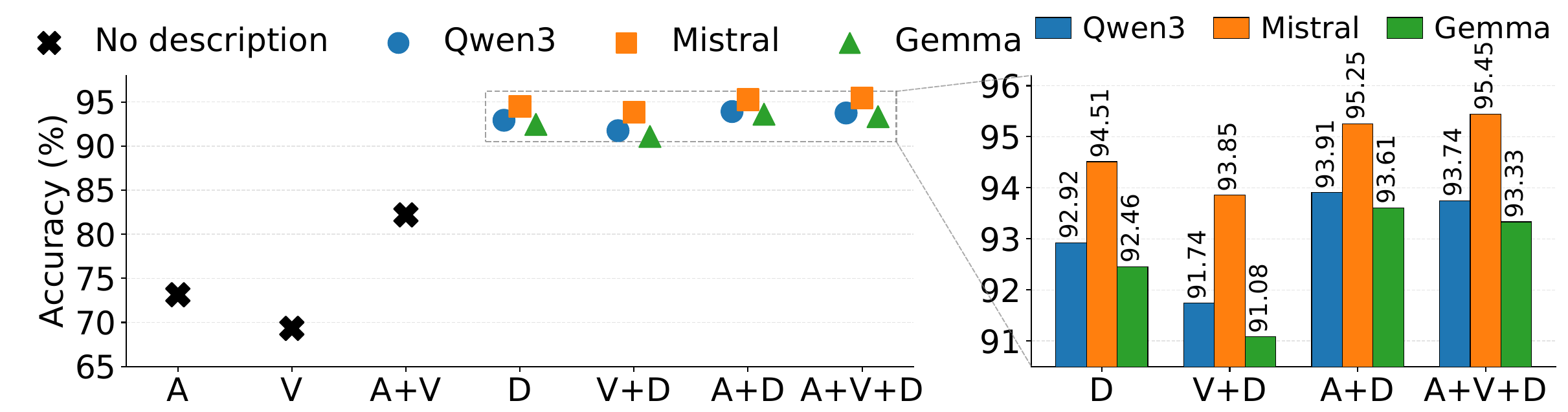}
    \vspace{-0.45cm}
    \caption{SVM-based urban scene classification accuracy for different combinations of audio (A), visual (V), and multimodal description (D) obtained using  Qwen3, Mistral, and Gemma models.}
    \label{fig: scene_classification}
\end{figure}

\begin{figure}[ht]
    \centering
    \includegraphics[scale=0.223]{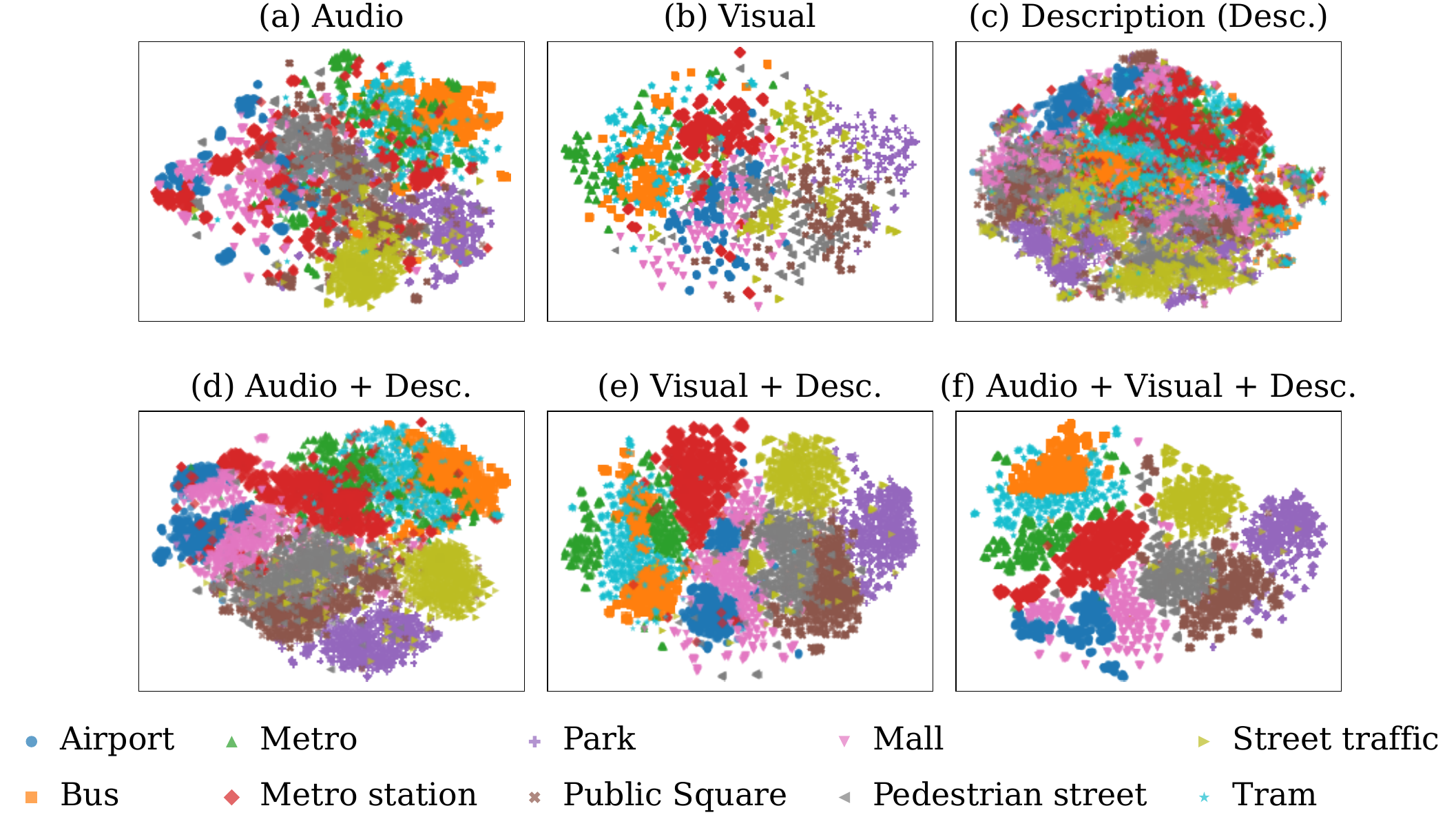}
    \vspace{-0.43cm}
    \caption{t-SNE visualizations of audio, visual, and Mistral-generated scene-description embeddings, and their multimodal combinations.
    }
    \label{fig: tsne}
\end{figure}

\noindent\textbf{Scene classification analysis:}
Figure~\ref{fig: scene_classification} presents the scene classification results. We first reproduce the audio, visual, and audio--visual baselines reported in \cite{wang2021curated}, with the reproduced audio--visual performance matching the reported result. 
The audio, visual, and audio--visual baselines achieve accuracies of 73.14\%, 69.33\%, and 82.19\%, respectively. 
The multimodal descriptions alone substantially outperform all three baselines, demonstrating that the generated descriptions preserve rich scene-discriminative information. Further improvements are obtained by combining the description embeddings with the audio and visual embeddings, as highlighted in the zoomed region of Fig.~\ref{fig: scene_classification}. These results indicate that the multimodal descriptions capture highly discriminative scene-level semantics. Although the description embeddings account for most of the classification performance, the audio and visual embeddings provide complementary information that leads to additional gains when fused with the multimodal descriptions.

The t-SNE visualizations in Fig.~\ref{fig: tsne} further show that incorporating multimodal descriptions leads to more distinct grouping of scene categories compared with the audio- and visual-only embeddings, hence qualitatively supporting the classification results.

\begin{figure}[t]
    \centering
    \includegraphics[scale=0.253]{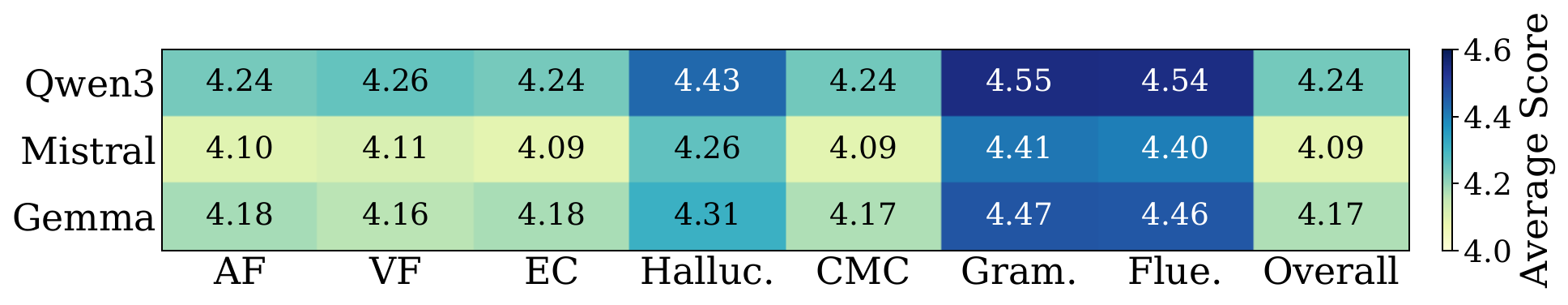}
    \vspace{-0.353cm}
    \caption{LLM-as-a-Judge evaluation of multimodal descriptions generated by different models. Scores are reported on a five point scale (higher is better). AF: audio fidelity; VF: visual fidelity; EC: event coverage; Halluc.: hallucination; CMC: cross-modal consistency; Gram.: grammar; Flue.: fluency.}
    \label{fig:LLM_evaluation}
\end{figure}

\noindent \textbf{Description quality analysis:} Fig.~\ref{fig:LLM_evaluation} presents the LLM-based evaluation results, indicating strong performance across all evaluation criteria and a low level of hallucination. Human evaluation of the Mistral-generated descriptions yields mean scores of 3.51, 4.14, 3.63, 3.22, 4.26, and 3.36 for audio fidelity, visual fidelity, event coverage, hallucination, fluency, and overall quality, respectively. The human ratings broadly support the LLM-based assessment, particularly for visual fidelity and fluency. While audio fidelity, event coverage, and hallucination receive comparatively lower scores, all criteria remain above the midpoint of the rating scale, suggesting that the generated descriptions provide generally accurate and informative representations of the audio--visual scenes.

\noindent \textbf{Ablation study:} We further evaluate instance-level cross-modal retrieval using CLAP and ImageBind embeddings. As shown in Table~\ref{tab:instance_retrieval}, retrieval performance is relatively low, particularly for text-to-audio retrieval, as the descriptions primarily capture scene-level semantics rather than clip-specific details. Moreover, clips from the same scene class often contain similar audio and visual content, making exact instance-level retrieval more challenging.


\begin{table}[t]
\centering
\caption{Instance-level cross-modal retrieval performance (\%).}
\label{tab:instance_retrieval}
\setlength{\tabcolsep}{2.7pt}
\LARGE
\renewcommand{\arraystretch}{1.05}
\resizebox{\columnwidth}{!}{%
\begin{tabular}{lccc|ccc|ccc}
\toprule
\multirow{2}{*}{\textbf{Description}}
& \multicolumn{3}{c}{\textbf{CLAP T$\rightarrow$A}}
& \multicolumn{3}{c}{\textbf{IMAGEBIND T$\rightarrow$A}}
& \multicolumn{3}{c}{\textbf{IMAGEBIND T$\rightarrow$V}} \\
& R@1 & R@5 & R@10
& R@1 & R@5 & R@10
& R@1 & R@5 & R@10 \\
\midrule



Multimodal (Qwen3)
& 0.24 & 0.83 & 1.42
& 0.07 & 0.39 & 0.81
& 3.91 & 15.15 & 24.77 \\

Multimodal (Mistral)
& 0.22 & 0.83 & 1.53
& 0.10 & 0.41 & 0.89
& 3.87 & 14.06 & 22.55 \\

Multimodal (Gemma)
& 0.24 & 0.91 & 1.57
& 0.11 & 0.41 & 0.73
& 3.02 & 11.83 & 19.35 \\

\bottomrule
\end{tabular}%
}
\end{table}

\begin{figure}[t]
    \centering
    \includegraphics[scale=0.2]{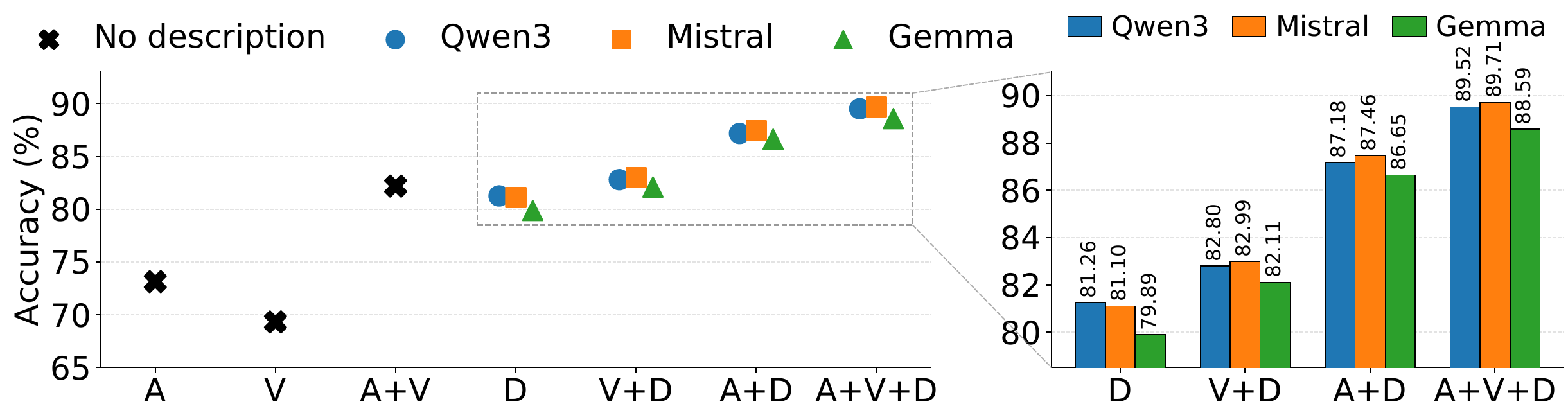}
    \vspace{-0.4cm}
    \caption{SVM-based scene classification accuracy without scene-labels in the instructions. A: Audio, V: Visual, D: Description.
    }
    \label{fig: scene_classification without scene label}
\end{figure}

\noindent \textbf{Effect of scene-label conditioning:} 
To investigate whether the multimodal descriptions primarily reflect scene-label information, we remove scene labels from the instructions ($I_A,I_V,I_{AV}$) and repeat the classification experiment. As shown in Figure~\ref{fig: scene_classification without scene label}, the descriptions remain strongly scene-discriminative, and combining them with audio or visual embeddings improves over the corresponding baselines. Although the label-guided pipeline yields higher overall performance, the no-label setting demonstrates that the generated descriptions retain discriminative scene-level semantics derived from the audio and visual content and provide complementary information beyond conventional audio--visual representations.

\section{Conclusion}

We introduced AVSD-Scenes, a dataset of 12,291 audio--visual scene descriptions generated from the TAU Urban Audio--Visual Scenes dataset. The proposed pipeline combines audio and visual descriptions using LLMs to generate unified multimodal scene descriptions. Experimental results show that the multimodal descriptions improve semantic alignment and cross-modal retrieval while providing informative scene-level representations for classification. The generated descriptions achieve up to 94.5\% scene classification accuracy using text embeddings alone, increasing to 95.4\% when combined with audio and visual embeddings. Human evaluation further supports the quality of the generated descriptions. Moreover, experiments without scene labels in the instructions demonstrate that the descriptions capture complementary semantic information beyond conventional audio--visual representations. Future work will focus on further scaling the dataset and improving the semantic grounding of multimodal descriptions by reducing hallucinations.

\bibliographystyle{IEEEbib}
\bibliography{strings,refs}

\end{document}